\documentclass{article}
\usepackage{spconf,amsmath,graphicx,hyperref}

\usepackage{colortbl} 
\usepackage{multirow}
\usepackage[table]{xcolor} 
\usepackage{amssymb} 
\usepackage[font=small,skip=2pt]{caption}

\usepackage{cite}
\usepackage{amsmath,amssymb,amsfonts}
\usepackage{graphicx}
\usepackage{textcomp}
\usepackage{xcolor}
\usepackage{hyperref}
\usepackage{indentfirst}
\usepackage{algorithm}
\usepackage{algpseudocode}
\usepackage{bbm}
\usepackage{booktabs}
\usepackage{siunitx}

\usepackage{soul}
\usepackage{lipsum}
\usepackage{hyperref}
\usepackage{cleveref}
\usepackage{amssymb}
\usepackage{marvosym}
\title{Enhancing Intent Understanding for Ambiguous Prompts: A Human-Machine Co-Adaption Strategy}
\name{%
\begin{tabular}{@{}c@{}}
Yangfan He$^{1*}$,
Jianhui Wang$^{2*}$,
Yijin Wang$^{3}$,
Yan Zhong$^{4}$,
Xinyuan Song$^{5}$,
Junjiang Lin$^{6}$,
Xinhang Yuan$^{7}$, \\
Jingqun Tang$^{8}$,
Yi Xin$^{9}$,
Hao Zhang$^{10}$,
Yuchen Li$^{11}$,
Zijian Zhang$^{12}$
Hongyang He$^{13}$,
Tianxiang Xu$^{21}$ \\
Miao Zhang$^{14}$,
Kuan Lu$^{15}$,
Menghao Huo$^{16}$,
Keqin Li$^{17}$,
Jiaqi Chen$^{18}$,
Tianyu Shi$^{19}$,
Jianyuan Ni$^{20\dagger}$ 
\end{tabular}
\thanks{* Equal contribution. † Corresponding author.}
}
\address{$^{1}$ UMN, $^{2}$ UESTC, $^{3}$ XDU, $^{4}$ NTU, $^{5}$ Emory, $^{6}$ Amazon, $^{7}$ WUSTL, $^{8}$ ByteDance,  $^{9}$ NJU,  $^{10}$ UCAS, \\ $^{11}$ Baidu, $^{12}$ Upenn, $^{13}$ UofWarwick, $^{14}$ THU, $^{15}$ Google, $^{16}$ SCU, $^{17}$ AMA, $^{18}$ Google $^{19}$ UofT, $^{20}$ JC, $^{21}$ PKU}
\begin{document}
%
\maketitle
\begin{abstract}
Current image generation systems produce high-quality images but struggle with ambiguous user prompts, making interpretation of actual user intentions difficult. Many users must modify their prompts several times to ensure the generated images meet their expectations. While some methods focus on enhancing prompts to make the generated images fit user needs, the model is still hard to understand users' real needs, especially for non-expert users. In this research, we aim to enhance the visual parameter-tuning process, making the model user-friendly for individuals without specialized knowledge and better understand user needs. We propose a human-machine co-adaption strategy using mutual information between the user's prompts and the pictures under modification as the optimizing target to make the system better adapt to user needs. We find that an improved model can reduce the necessity for multiple rounds of adjustments. We also collect multi-round dialogue datasets with prompts and images pairs and user intent. Various experiments demonstrate the effectiveness of the proposed method in our proposed dataset. Our dataset and annotation tools will be available.
\end{abstract}
\section{Introduction}
\label{sec:intro}

\begin{figure}[t!]
    \centering
    \includegraphics[width=1\linewidth]{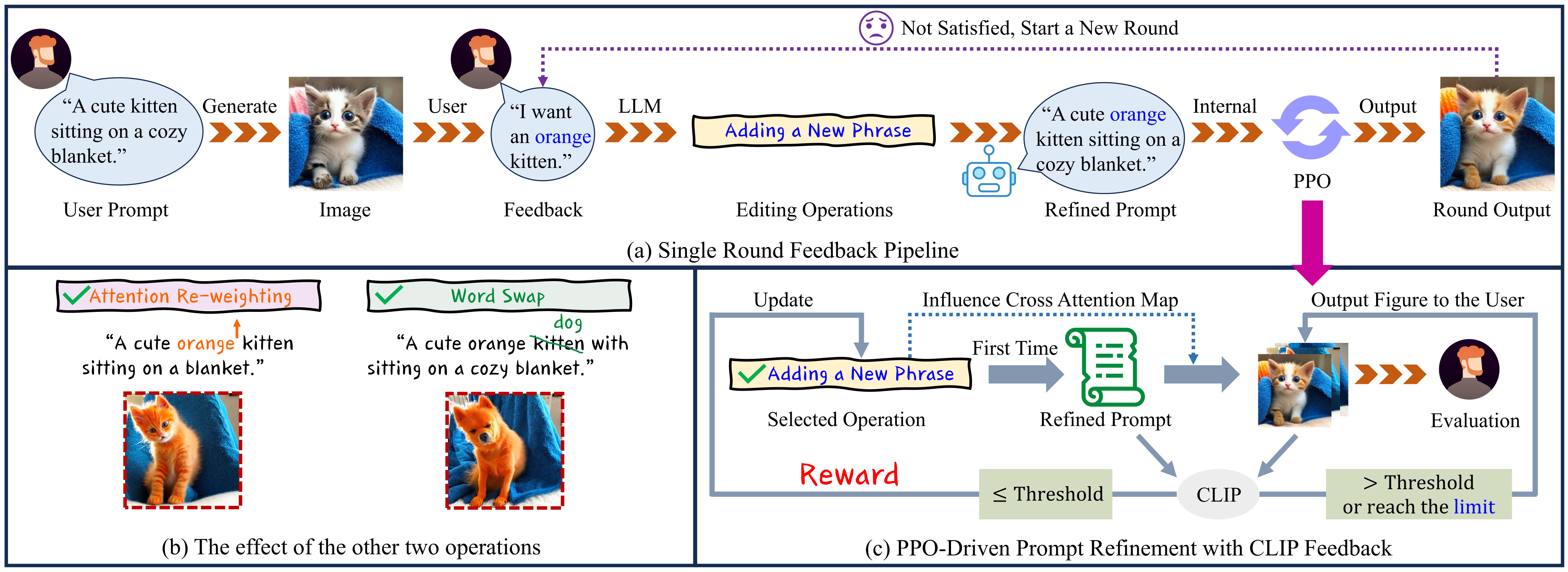}
    \caption{A visual illustration of the multi-round human–machine collaborative image generation and refinement process, incorporating phrase addition, word substitution, and attention re-weighting operations within a PPO-based reinforcement learning loop guided by CLIP-based reward feedback.}
    \label{fig:prompt_refinement_pipeline}
\end{figure}

Generative AI shows promise for streamlining creative and non-creative tasks. Models like DALLE 3 \cite{betker2023improving}, Imagen~\cite{saharia2022photorealistictexttoimagediffusionmodels}, Stable Diffusion~\cite{rombach2022high, esser2024scalingrectifiedflowtransformers}, and Cogview 3~\cite{zheng2024cogview3finerfastertexttoimage} excel at generating lifelike images from text~\cite{gozalo2023chatgpt,croitoru2023diffusionSurvey,gu2023matryoshka,xing2023survey}. However, challenges remain in producing higher-resolution images, improving semantic alignment, and enhancing user-friendly interfaces~\cite{frolov2021adversarial,wang2023imageneditor,mehrabi2023resolving,liang2023rich}, as models often struggle with nuanced human instructions. Additionally, the impact of variable adjustments on the final image is not always clear, posing challenges for non-expert users who haven't systematically studied prompt engineering. Unlike traditional models that require a deep understanding of underlying mechanisms and control elements, our approach enables users to adjust and optimize image generation with minimal technical knowledge. Inspired by human-in-the-loop co-adaptation~\cite{reddy2022first}, our model evolves with user feedback to better meet user expectations. Figure~\ref{fig:prompt_refinement_pipeline} illustrates the operational flow as interacted by users.

We contribute: (i) We propose visual co-adaptation (VCA), an adaptive framework that fine-tunes user prompts using a pre-trained language model enhanced through reinforcement learning, aligning image outputs more closely with user preferences and creating images that truly reflect individual styles and intentions. (ii)  Our work considers incorporating human feedback within the training loops of diffusion models. By assessing its impact, we demonstrate how human-in-the-loop methods can surpass traditional reinforcement learning in enhancing model performance and output quality. (iii) Through comparative analysis, we explore the superiority of mutual information maximization over conventional reinforcement learning in tuning model outputs to user preferences. Additionally, we introduce an interactive tool that grants non-experts easy access to advanced generative models, enabling the creation of personalized, high-quality images, thus broadening the applicability of text-to-image technologies in creative domains.
\vspace{-0.3 cm}
\section{Method}
\subsection{Mutual Information for Co-Adaptation}
Our system adapts to user preferences by maximizing the mutual information \( I(\mathcal{X}; \mathcal{Y}) \) between user inputs (prompts and feedback) and generated images, refining its output each round to better align with the user’s intent. Mutual information \( I(\mathcal{X}; \mathcal{Y}) \) measures the dependency between the user input \( \mathcal{X} \) and the generated image \( \mathcal{Y} \):
\begin{equation}
\small
I(\mathcal{X}; \mathcal{Y}) = \int_{x} \int_{y} p(x, y) \log \frac{p(x, y)}{p(x)p(y)} \, dy \, dx,
\end{equation}
where \( p(x, y) \) is the joint probability distribution of \( x \) and \( y \), while \( p(x) \) and \( p(y) \) are the marginal distributions. 
\begin{align}
\small
I(\mathcal{X}; \mathcal{Y}) = \frac{1}{2}\log\Big( \frac{\det(\Sigma_{\mathcal{X}}) \det(\Sigma_{\mathcal{Y}})}{\det(\Sigma_{\mathcal{Z}})} \Big),
\end{align}
with \( \Sigma_{\mathcal{X}} \) and \( \Sigma_{\mathcal{Y}} \in \mathbb{R}^{D \times D} \) as the covariances of \( \mathcal{X} \) and \( \mathcal{Y} \), respectively, and \( \Sigma_{\mathcal{Z}} \in \mathbb{R}^{2D \times 2D} \) as the covariance matrix of the concatenated vector \( \mathbf{z} = [\mathcal{X}; \mathcal{Y}] \) representing the joint distribution. For high-dimensional data, we encode \( \mathcal{X} \) and \( \mathcal{Y} \) using CLIP encoders~\cite{radford2021learning,xu2024imagereward,wu2023better}, maximizing \( I(\tilde{\mathcal{X}}; \tilde{\mathcal{Y}}) \). In practice, we approximate mutual information by directly optimizing the CLIP score, which effectively captures semantic alignment between generated images and user inputs. This alignment score serves as a surrogate objective, guiding training to dynamically adapt to user feedback without explicitly computing mutual information. This approach builds on retrieval-augmented generation techniques~\cite{lewis2020retrieval,pan2023automatically,huang2022large} for better alignment with user preferences.

\noindent\textbf{Adaptive Feedback Loop.} The feedback loop updates model parameters \( \theta \) based on user feedback \( f \), refining predictions over time:
\begin{equation}
\small
\theta_{\text{new}} = \theta_{\text{old}} - \eta \nabla I(\mathcal{X}; \mathcal{Y} \mid f),
\end{equation}
where \( \eta \) is the learning rate. User feedback continuously enhances the relevance and accuracy of generated images. For details on reward-based optimization using CLIP scores, refer to Section~\ref{sec:clip}.

\subsection{Editing Operations}
To facilitate co-adaptation and maximize mutual information, the system iteratively refines the prompt and adjusts attention maps $M_t$, alignment parameters $A_t$, and scaling factors $c_t$ through gradient ascent within an RL framework~\cite{hertz2022prompt,avrahami2023blended,nichol2021glide}, using the CLIP similarity score as a reward signal to align generated images with user intent.
\vspace{-0.2 cm}
\subsubsection{Word Swap.}
The Word Swap operation lets users replace tokens in the prompt $P_t = \{w_1, w_2, \dots, w_n\}$ to modify key attributes of the generated image, enabling instruction-based editing~\cite{brooks2023instructpix2pix,wang2023imageneditor,gafni2022make}, such as changing $\mathcal{P}=$``a serene blue lake'' to $\mathcal{P}^*=$``a vibrant green forest''. By replacing $w_i$ with $w_i'$, the prompt updates to $P_{t+1} = \{w_1, \dots, w_i', \dots, w_n\}$, and the attention map $M_t$ is conditionally updated to preserve the overall composition:
\begin{equation}
\small
Edit(M_t, M_t^*, t) := 
\begin{cases}
M_t^*, & \text{if } t < \tau, \\
M_t, & \text{otherwise}.
\end{cases}
\end{equation}
Here, $\tau$ controls the number of steps for injecting the updated attention map $M_t^*$, which is refined through gradient ascent to optimize the image based on user feedback without disrupting its structure.
\begin{equation}
\small
M_t^* = M_t^* + \eta \nabla_{M_t^*} \mathcal{R}(M_t^*),
\end{equation}
where $\eta$ is the learning rate, and $\mathcal{R}(M_t^*)$ is the reward  used to align the attention map with user preferences.
\vspace{-0.2 cm}
\subsubsection{Adding a New Phrase.}
The Adding a New Phrase operation lets users introduce new elements into the generated image by inserting tokens into the prompt, such as changing $\mathcal{P}=$``a tranquil garden'' to $\mathcal{P}^*=$``a tranquil garden with blooming flowers''. The updated prompt becomes $P_{t+1} = \{w_1, \dots, w_{i}, w_{\text{new}}, w_{i+1}, \dots, w_n\}$, and coherence is maintained by adjusting attention maps with an alignment function $A(j)$ that maps indices from $M_{t+1}$ to the original $M_t$:
\begin{equation}
\small
\resizebox{\linewidth}{!}{$
\left( \text{Edit} \left( M_t, M_{t}^*, t \right) \right)_{i,j} := 
\begin{cases}
    (M_t^*)_{i,j}, & \text{if } A(j) = \text{None}, \\
    (M_t)_{i,A(j)}, & \text{otherwise}.
\end{cases}
$}
\end{equation}
The adjustment ensures existing tokens retain their attention distribution while new tokens receive appropriate attention, with the alignment function $A_t$ updated through gradient ascent to maintain consistency with user feedback.
\begin{equation}
\small
A_t = A_t + \eta \nabla_{A_t} \mathcal{R}(A_t),
\end{equation}
where $\eta$ is the learning rate and $\mathcal{R}(A_t)$ is the reward function.
\vspace{-0.5 cm}
\subsubsection{Attention Re-weighting.}
In the Attention Re-weighting method, token influence is adjusted to enhance or diminish features. For instance, scaling the attention map for a specific token (e.g., “sunflower”) with a parameter $c \in [-2, 2]$ allows users to control its prominence in the image:
\begin{equation}
\small
\resizebox{\linewidth}{!}{$
\left( \text{Edit} \left( M_{t}, M_{t+1}, t \right) \right)_{i,j} := 
\begin{cases}
c \cdot M_t(i,j), & \text{if } j = j^*, \\
M_t(i,j), & \text{otherwise}.
\end{cases}
$}
\end{equation}
This parameter $c$ provides intuitive control over the induced effect. The scaling parameter $c_t$ is updated as follows:
\begin{equation}
\small
c_t = c_t + \eta \nabla_{c_t} \mathcal{R}(c_t),
\end{equation}
Here, $\eta$ is the learning rate and $\mathcal{R}(c_t)$ represents the reward function, which guides gradient ascent to align generation with user preferences.

\begin{table}[t]
\centering
\small 
\caption{Comparative analysis of prompt refinement from 100 users, with metrics scored on a 0-5 scale and response time reflecting the average duration for self-reflection and multi-dialogue processes.}
\label{tab:prompt_comparison}
\resizebox{\linewidth}{!}{
\begin{tabular}{@{}l S[table-format=1.1] S[table-format=1.1]@{}}
\toprule
\multirow{2}{*}{\textbf{Metric \& Category}} & \multicolumn{2}{c}{\textbf{Refinement Type}} \\
\cmidrule(l){2-3}
& \textbf{Self-reflection} & \textbf{Multi-dialogue} \\
\midrule
\textbf{Prompt Quality} & & \\
\quad Clarity & 4.23 & \textbf{4.75} \\
\quad Detail Level & 4.11 & \textbf{4.28} \\
\quad Purpose Adaptability & 3.30 & \textbf{4.82} \\
\midrule
\textbf{Image Reception} & & \\
\quad User Satisfaction & 3.02 & \textbf{4.70} \\
\quad CLIP Score & 0.38 & \textbf{0.45} \\
\quad SSIM Score & 0.72 & \textbf{0.84} \\
\quad LPIPS Score & 0.24 & \textbf{0.13} \\
\quad ImageReward & 0.22 & \textbf{0.32} \\
\midrule
\textbf{Response Time} & \textbf{3.4 s} & 12.7 s \\
\bottomrule
\end{tabular}
}
\vspace{-0.5 cm}
\end{table}

\subsection{TD Error Experience Replay with Joint Training}
\label{sec:clip}
Our model uses PPO to update its policy $\pi_\theta(a_t \mid s_t)$ based on user feedback~\cite{lee2023aligning,wu2023better}, where the state $s_t$ includes the image and fixed prompt with rewritten feedback from the previous round, and the action $a_t$ governs image generation or editing to optimize the cumulative reward:
\begin{equation}
\small
J(\theta) = \mathbb{E}_{\pi_\theta} \left[ \sum_{t=0}^T \gamma^t r_t \right],
\end{equation}
With \(\gamma \in [0,1]\) as the discount factor and \(T\) as the time horizon, \(V(s_i, \theta)\) estimates the expected cumulative reward, while \(r_i = \text{CLIP}(I_i, P_i)\) measures the immediate reward. The value function $V^\pi(s_t)$ is updated by minimizing the temporal difference (TD) error:
\begin{equation}
\small
\delta_t = r_t + \gamma V(s_{t+1}, \theta) - V(s_t, \theta).
\end{equation} 
Prioritized experience replay uses non-uniform sampling, storing transitions \((s_r, a_t, r_t, s_{t+1})\) and TD errors in a memory pool \(\mathcal{M}\), prioritizing unfamiliar states and recent trajectories when minimizing TD errors based on a forgetting curve and historical data:
\begin{equation}
\small
p_i \propto e^{-\lambda t_i} \cdot |\delta_i| + \epsilon,
\end{equation}
The sampling probability \(p_i\) is proportional to \(e^{-\lambda t_i} \cdot |\delta_i| + \epsilon\), with \(\lambda\) controlling the forgetting rate, \(t_i\) the experience age, \(|\delta_i|\) the TD error, and \(\epsilon\) ensuring non-zero probability; learning rates are adjusted by scaling each experience by \((n \cdot p_t)^{-\beta}\), where \(\beta\) increases from a small value to 1, and PPO updates the policy model (Three strategies of editing attention) by maximizing the objective function:
\begin{equation}
\small
\resizebox{.9\linewidth}{!}{$
\mathcal{L}^{\text{PPO}}(\theta) = \mathbb{E}_t \left[
\begin{aligned}
&\min \left( \rho_t(\theta) \hat{A}_t, \right. \\
&\quad \text{clip}\left( \rho_t(\theta), 1 - \epsilon, 1 + \epsilon \right) \hat{A}_t \Big)
\end{aligned}
\right]
$}
\end{equation}
The policy, consisting of three cross-attention map editing strategies, is refined by jointly maximizing rewards through gradient ascent with learning rate \(\alpha_r\) and minimizing TD errors via gradient descent with learning rate \(\alpha_\delta\), where \(\rho_t(\theta) = \frac{\pi_\theta(a_t \mid s_t)}{\pi_{\theta_{\text{old}}}(a_t \mid s_t)}\) is the probability ratio between new and old policies, and the advantage function \(\hat{A}_t = r_t + \gamma V^\pi(s_{t+1}) - V^\pi(s_t)\) reflects the relative improvement of the current action. At each iteration $t$ in PPO Training, we compute the CLIP similarity score: $s_t = \text{CLIP}(I_t, P_t)$. If $s_t \geq \tau$, we return $I_t$ to the user; otherwise, we continue editing until $t = N_{\text{max}}$, then return $I_t$.

\begin{table}[t]
\centering
\small
\caption{Comparing the effects of using PPO-based reinforcement learning (RL) versus using only three editing strategies (without RL) on user satisfaction and image quality metrics. Data is averaged from 100 diverse users, with final CLIP and Aesthetic Scores based on 0-5 subjective ratings.}
\label{tab:RL}
\resizebox{\linewidth}{!}{  
    \begin{tabular}{lcc}
    \toprule
    \textbf{Metrics} & \textbf{With PPO (RL)}  & \textbf{Without PPO (Only 3 Strategies)}  \\ 
    \midrule
    \textbf{Rounds} & \textbf{4.3} & 6.9 \\
    \textbf{User Satisfaction} & \textbf{4.70} & 4.14 \\
    \textbf{CLIP Score} & \textbf{0.45} & 0.28 \\
    \textbf{SSIM Score} & \textbf{0.84} & 0.78 \\
    \textbf{LPIPS Score} & \textbf{0.13} & 0.18 \\
    \textbf{ImageReward} & \textbf{0.32} & 0.26 \\
    \bottomrule
    \end{tabular}
}
\vspace{-0.3 cm}
\end{table}

\begin{table}[t]    \centering
    \small  
    \caption{Assessing cross attention (CA) impact by averaging CLIP and Aesthetic Scores from final interactions, based on 0-5 user ratings of image attractiveness, using data from a diverse sample of 100 users.}
    \label{tab:CA}
    
    \begin{tabular}{lcc}
    \toprule
    Metrics & Edited CA & Normal CA \\
    \midrule
    \textbf{Rounds} & \textbf{3.7} & 6.1 \\
    \textbf{User Satisfaction} & \textbf{4.70} & 3.94 \\
    \textbf{CLIP Score} & \textbf{0.45} & 0.41 \\
    \textbf{SSIM Score} & \textbf{0.84} & 0.81 \\
    \textbf{LPIPS Score} & \textbf{0.13} & 0.81 \\
    \textbf{ImageReward} & \textbf{0.32} & 4.48 \\
    \bottomrule
    \end{tabular}
    \vspace{-10pt}
\end{table}

\begin{figure*}[t]
    \centering
    \includegraphics[width=0.95\textwidth]{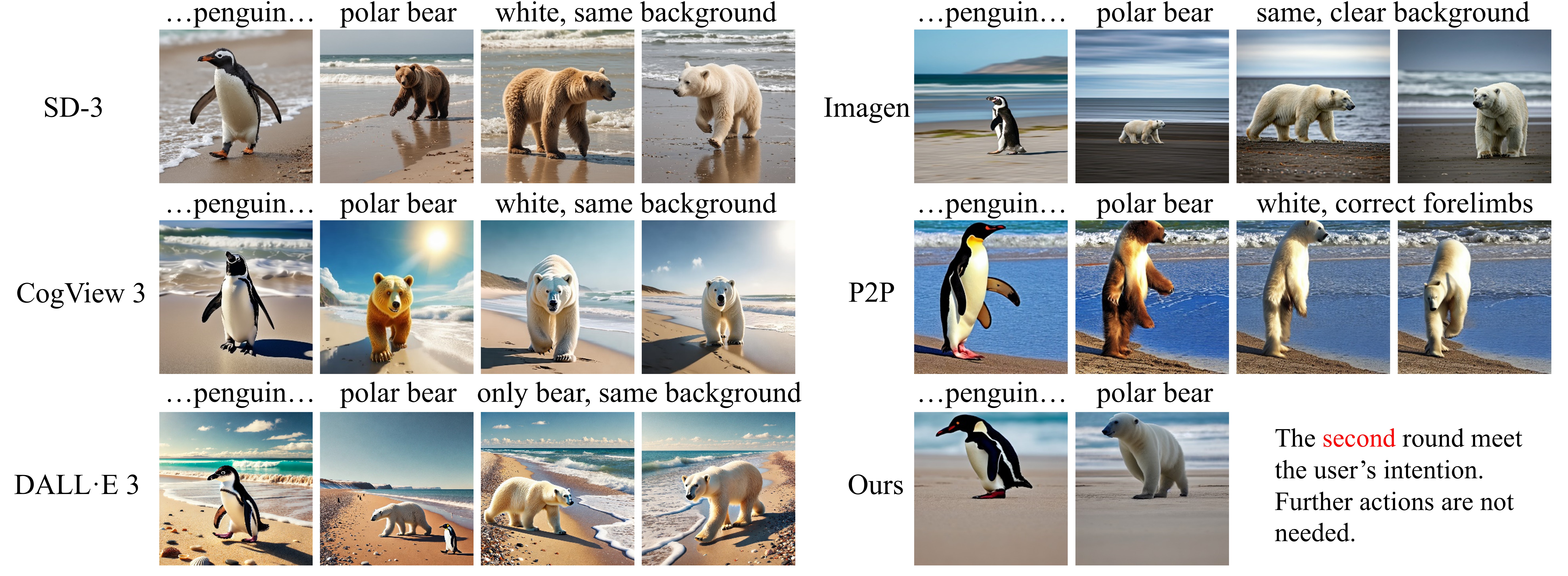} 
    \caption{Visualization result comparisions. } 
    \label{fig:vis1}
    \vspace{-0.3 cm}
\end{figure*}

\vspace{-0.2 cm}
\section{Experiments}

\noindent\textbf{Dataset.} We developed a Q\&A tool to generate 1,673 multi-turn dialogue annotations stored in JSON files, each containing prompts, Q\&A sequences, image paths, and human ratings. Thirty diverse annotators rated image relevance, coherence, and artistic quality, with evaluations averaged for consistency. Our training set consists of 1673 JSON files with prompts, Q\&A sequences, image paths, unique IDs, and ratings for image alignment and fidelity. This dataset informs our model on user expectations, analyzing subjects, emotions, styles, and settings. Feedback refines prompts, helping the model capture complex artistic intentions. We use 95\% of the data for training and 5\% for validation, supporting few-shot learning for better performance and user satisfaction.

\noindent\textbf{Experiment Settings.} We train the model over 12,000 episodes using 4 NVIDIA 4090 GPUs with Proximal Policy Optimization (PPO)~\cite{schulman2017proximal}, initialized from a fine-tuned baseline. Each batch (size 256) runs 4 PPO epochs with a learning rate of \(5 \times 10^{-5}\), and the value and KL reward coefficients are meticulously calibrated to 2.2 and 0.3, respectively. We use diverse beam search~\cite{vijayakumar2016diverse} (beam size 8, diversity 1.0) for diverse generation, with lengths sampled between 15–75. One image is generated per prompt, using CLIP score as reward. We evaluate using standard metrics including FID~\cite{heusel2017FID} and LPIPS~\cite{zhang2018unreasonable}. Human-written prompts (via ChatGPT) reflect intent with minimal structural change. 

\noindent\textbf{Prompt Refinement.}
Table~\ref{tab:prompt_comparison} compares self-reflection and multi-round dialogue refinements. Self-reflection is faster (3.4s vs. 12s), but multi-round dialogue yields higher user satisfaction (4.7 vs. 3.0), better Purpose Adaptability (4.8 vs. 3.3), and marginal gains in Clarity (4.7 vs. 4.2) and Detail (4.2 vs. 4.1). 

\begin{table}[t]
\small
    \renewcommand{\arraystretch}{1.1}
    \centering
    \caption{Model performance in three tasks measured by SSIM. Higher SSIM indicates better performance.}
    \label{tab:ssim_performance}
    \resizebox{\linewidth}{!}{  
    \begin{tabular}{l cc c cc c cc c | cc}
    \toprule
    \multirow{2}{*}{Model} & \multicolumn{2}{c}{Task 1} && \multicolumn{2}{c}{Task 2} && \multicolumn{2}{c}{Task 3} && \multicolumn{2}{c}{Overall} \\
    \cmidrule{2-3} \cmidrule{5-6} \cmidrule{8-9} \cmidrule{11-12}
    & Mean & SD && Mean & SD && Mean & SD && Mean & SD \\
    \midrule
    CogView 3  & $0.382$ & $0.017$ && $0.330$ & $0.079$ && $0.207$ & $0.093$ && $0.306$ & $0.102$ \\
    SD 3    & $0.446$ & $0.043$ && $0.437$ & $0.091$ && $0.334$ & $0.094$ && $0.406$ & $0.077$ \\
    Imagen  & $0.415$ & $0.008$ && $0.388$ & $0.017$ && $0.296$ & $0.044$ && $0.366$ & $0.053$ \\
    P2P      & $0.510$ & $0.007$ && $0.289$ & $0.005$ && $0.107$ & $0.011$ && $0.302$ & $0.165$ \\
    DALL·E 3                       & $0.537$ & $0.007$ && $0.297$ & $0.025$ && $0.367$ & $0.014$ && $0.400$ & $0.102$ \\
    Ours                                                        & $0.623$ & $0.004$ && $0.468$ & $0.007$ && $0.496$ & $0.003$ && $0.561$ & $0.054$ \\
    \bottomrule
    \end{tabular}
    }
    \vspace{-0.5 cm}
\end{table}

\noindent\textbf{Task-Based Model Performance Evaluation.}
Table~\ref{tab:ssim_performance} compares the SSIM performance of six models across three tasks evaluated by 81 users, who generated images aligned with target images through prompts. Task 1 uses simple targets, Task 2 more complex ones, and Task 3 highly intricate and detailed images. Lower standard deviation (SD) in Task 1 indicates more consistent results, while higher SD in Task 3 reflects increased performance variation due to complexity.

\begin{table}[t]
\centering
\small
\caption{Impact of Removing Each Strategy on Model Performance. W.S denotes Word Swap.A.P D denotes Adding Phrase. A.R denotes Attention Re-weighting}
\label{tab:ablation_strategies}
\resizebox{\columnwidth}{!}{
\begin{tabular}{lcccc}
\toprule
\textbf{Metric} & \textbf{w/o W.S} & \textbf{w/o A.P} & \textbf{w/o A.R} & \textbf{Ours} \\
\midrule
\textbf{Dialogue Rounds}    & 5.8        & 6.2         & 5.5                      & \textbf{4.3}      \\
\textbf{User Satisfaction}  & 4.20/5     & 4.10/5      & 4.50/5                   & \textbf{4.73/5}   \\
\textbf{CLIP Score}         & 0.85       & 0.80        & 0.88                     & \textbf{0.92}     \\
\textbf{Aesthetic Score}    & 4.70/5     & 4.65/5      & 4.80/5                   & \textbf{4.89/5}   \\
\bottomrule
\end{tabular}
}
\vspace{-0.5 cm}
\end{table}

\noindent\textbf{Ablation Study.} As shown in Table~\ref{tab:RL}, Reinforcement Learning (RL) tuning reduces dialogue rounds (4.3 vs. 6.9), improves CLIP scores (0.92 vs. 0.83), and increases user satisfaction (4.73 vs. 4.14), while maintaining comparable aesthetic quality (4.89 vs. 4.88), owing to better functionality and feedback adaptation. Table~\ref{tab:CA} further demonstrates that edited cross attention (CA) dynamically adapts to context and feedback, achieving fewer rounds (3.7 vs. 6.1), higher CLIP scores (0.88 vs. 0.81), improved satisfaction (4.82 vs. 3.94), and enhanced aesthetics (4.71 vs. 4.48). Moreover, Table~\ref{tab:ablation_strategies} indicates that removing strategies like Word Swap and Attention Re-weighting increases dialogue rounds (up to 6.2) and lowers satisfaction (to 4.10/5), whereas the full model performs best with 4.3 rounds, a 0.92 CLIP score, and 4.73/5 satisfaction. Finally, Figure~\ref{fig:vis1} exemplifies efficient user intention alignment, such as adding croutons to soup images, achieving full capture by the third round with minimal edits.

\vspace{-0.2 cm}
\section{Conclusion}
We propose a novel human–machine co-adaptation framework for iterative text-to-image generation that enhances user–model alignment through Word Swap, Phrase Addition, and Attention Re-weighting strategies within a PPO-based reinforcement learning loop using CLIP reward. Experiments demonstrate its consistent superiority over self-refinement and single-strategy baselines across user satisfaction, CLIP, SSIM, LPIPS, confirming strong practical value.

\vfill\pagebreak
\footnotesize
\bibliographystyle{IEEEbib}
\bibliography{strings,refs}

\end{document}